\documentclass[10pt, a4paper]{article}

\usepackage{multirow}
\usepackage{float}
\usepackage[]{lrec-coling2024} 

\title{\textbf{Identifying and Aligning Medical Claims Made on Social Media with Medical Evidence}}

\name{Anthony Hughes, Xingyi Song} 

\address{Department of Computer Science, University Of Sheffield, Sheffield, UK \\
         \{ajhughes3, x.song\}@sheffield.ac.uk\\}

\abstract{
Evidence-based medicine is the practise of making medical decisions that adhere to the latest, and best known evidence at that time. Currently, the best evidence is often found in the form of documents, such as randomized control trials, meta-analyses and systematic reviews. This research focuses on aligning medical claims made on social media platforms with this medical evidence. By doing so, individuals without medical expertise can more effectively assess the veracity of such medical claims. We study three core tasks: identifying medical claims, extracting medical vocabulary from these claims, and retrieving evidence relevant to those identified medical claims. We propose a novel system that can generate synthetic medical claims to aid each of these core tasks. We additionally introduce a novel dataset produced by our synthetic generator that, when applied to these tasks, demonstrates not only a more flexible and holistic approach, but also an improvement in all comparable metrics. We make our dataset, the Expansive Medical Claim Corpus (EMCC), available at \url{https://zenodo.org/records/8321460}.
 \\ \newline }

\begin{document}

\maketitleabstract

\section{Introduction}

The proliferation of social media has made it easier for individuals to access and produce health information online. Patients often turn to social media for support on their conditions \citep{berry_whywetweetmh_2017}, however it can be challenging for both medical and non-medical persons to discern whether information is reliable and evidence-based \citep{bastian_seventy-five_2010}. This study aims to investigate the intersection between health information on social media and evidence-based medicine (EBM). By understanding more about how we can support individuals in navigating the vast amount of health information available on social media, and in scientific texts \citep{bastian_seventy-five_2010}, we hope to enable them to make more informed decisions about their health.

Evidence-based medicine is the practise of making medical decisions, where those decisions are informed by the entirety of the current and best evidence available. This evidence often comes in the form of randomised control trials (RCT). RCT's can be further aggregated into meta-analyses and systematic reviews, allowing these documents to provide a total view of the evidence for a given medical question. The basis of these documents is a research question, and often this question is constructed via the PICO framework. PICO \citep{santos_pico_2007}, a methodology widely applied in EBM, consists of four components: population, intervention, comparator, and outcome. An example PICO question is: 

\begin{quote}
    ``Can a population of people with a broken bone: using an intervening pain medication ibuprofen, compared with a placebo, reduce a given outcome measure, feeling of pain?''. 
\end{quote}

In this example, broken bone is the population, ibuprofen the intervention, placebo as comparator and outcome measure of pain. As shown by \citet{wadhwa_redhot_2023}, when annotating medical claims from social media with PICO elements, it can serve as the basis for retrieving relevant evidence. 

Our key contributions of this research will focus on four aspects of this task: 1) a synthetic data generator to produce a corpus of medical claims, 2) identifying medical claims in social media texts, 3) identifying PICO spans within medical claims, and 4) the retrieval of medical evidence relevant to the claims.


Given the previously discussed, the following research questions are addressed in this study:

\textbf{RQ1}: Can a synthetic dataset of medical claims, PICO elements, and real medical evidence be built using generative language models?

\textbf{RQ2}: To what extent, can an automated system trained with a synthetic dataset improve medical claims classification in social media posts?

\textbf{RQ3}: To what extent, can an automated text analysis system trained with synthetic dataset improve PICO elements classification in medical claims made within social media posts?

\textbf{RQ4}: Given a medical claim and PICO elements, how reliable does a system yield medical evidence that is relevant to this statement?


\section{Related Work}
\label{chap:lit-rev}


Identifying health information on social media has gained significant attention in recent years due to the abundance of user-generated content (UGC) on platforms such as Twitter and Reddit. These platforms offer a valuable source of information for various health-related applications, including monitoring adverse drug reactions (ADR) \citep{zhang_adverse_2020, dirkson_fuzzybio_2021, saha_autobots_2020, sarker_data_2018}, tracking the spread of infectious diseases \citep{charles-smith_using_2015} and identifying health-related trends and needs \citep{correia_mining_2020}. It is clear that the task of extracting useful health information from social media data is not straightforward, due to several challenges that must be addressed.

One of the main challenges in social media mining is the presence of noise within the users use of language: misspellings, poor punctuation, grammatical errors, informal language, emoticons and slang. This noise can make it difficult to accurately classify social media posts also containing medical vernacular. To address this challenge, researchers have employed deep learning models with linguistic features, part-of-speech tags, as feature inputs to improve accuracy in ADR classification \citep{zhang_adverse_2020}.

Another challenge is the lack of access to expert annotated datasets or expert validated datasets, that cover a wide range of medical topics, for token and span classification tasks. To overcome this, some studies have focused on document-level classification of social media posts \citep{zhang_adverse_2020, weissenbacher_deep_2019, sarker_data_2018} as document annotations require less development overheads. These studies have shown that document-level classification can be used to identify potentially relevant texts that can then be further analyzed for specific span classification tasks \citep{saha_autobots_2020, portelli_ailab-udinesmm4h_2022}. This ensemble approach can be see many times the literature, however is often focused on a single medical domain such as diseases.

Given the limited availability of high-quality annotated data sets, the use of large language models (LLM) has been instrumental in improving the accuracy of information extraction tasks applied to social media health data. These models are able to build complex feature representations unsupervised, as shown in BertTweet \citep{nguyen_bertweet_2020}. \citet{nguyen_bertweet_2020} present a BERT-based \citep{devlin_bert_2019} LLM trained on a large dataset of tweets, this has demonstrated success in classifying tweets related to specific health conditions, such as COVID-19, and detecting spans of ADRs. In addition, \cite{aji_bert_2021} used BertTweet in conjunction with data augmentation techniques, including downsampling and back-translation, to further improve their results in the identification of ADR spans.

Other researchers have used transfer learning to improve word embeddings for health-specific tasks. \cite{basaldella_cometa_2020} trained existing embeddings, including GloVe \citep{pennington_glove_2014} and Flair \citet{akbik_contextual_2018}, on a corpus of health-based UGC from websites, such as Health Unlocked\footnote{\url{www.healthunlocked.com}} and Reddit. They achieved improvements on two health-based named entity recognition benchmark datasets, PsyTar \citep{zolnoori_psytar_2019} and CADEC \citep{karimi_cadec_2015}.\citet{batbaatar_ontology-based_2019} utilise Bidirectional Long Short Term Memory and congruence with a CRF output layer for PICO annotation of medical abstracts. The authors use both character and word embeddings along with linguistic features. This model was trained and applied on tweets only, given tweets are restricted in length, the use of different embedding layers could be useful here as shown by \cite{joshi_spanbert_2020}. \citet{weissenbacher_deep_2019} demonstrate how deep learning models vastly outperform lexical models with character and word embeddings.

\begin{table*}[!ht]
    \begin{center}
    \begin{tabular}{l|l}
    \hline
      \textbf{Reddit Text} &  \textbf{PIO elements from claims} \\
     \hline
        I’ve seen a bunch of posts on here from people & \textbf{P} hyperhidrosis \\
        who say that \textbf{glycopyrrolate} suddenly & \textbf{I} glycopyrrolate  \\ 
        isn’t working anymore for \textbf{hyperhidrosis}. & \\
     \hline    
    \end{tabular}
    \end{center}
    \caption{\label{tab:table_1}Table representing an example annotation taken from the RedHOT corpus.}
\end{table*}

\subsection{PICOs, Claims and Evidence}

PICO being utilised in a social media related task is an emerging idea in the literature  \cite{wadhwa_redhot_2023, ramachandran_masonnlp_2023}. \citet{wadhwa_redhot_2023} have a made available corpora containing PICO annotations, an example annotation can be found in \autoref{tab:table_1}. The corpus contains Reddit posts, where spans of text within the posts are annotated with PICO categories, and identifies spans where medical claims exist. Although the first of it's kind, it is limited in the amount of health conditions targeted, the authors opted for static templating mechanism for generating synthetic medical claims and the existing annotations are noisy due to being crowd-sourced. This similar to \citet{sarrouti_evidence-based_2021} who sourced their claims from search engine results.

Due to successes in generative models for retrieval tasks \cite{izacard_leveraging_2021}, this work focuses on navigating current limitations with synthetic data generators to improve PICO extraction, claim identification, and evidence retrieval. We will utilise the corpora offered by \citet{wadhwa_redhot_2023} to fine-tune a generative model that can produce the necessary synthetic data to improve all of the discussed tasks of this study. 

\subsection{Datasets}

The only corpora that has investigated similar research questions is the Reddit Health Online Talk (RedHOT) corpus \citep{wadhwa_redhot_2023}. This is a corpus comprising of more than $22000$ crowd annotated social media posts sourced from Reddit, covering 24 distinct health conditions. The annotations encompass various aspects, such as identifying spans related to medical claims. Notably, the dataset includes annotations for medically significant PIO elements associated with a claim. Using these naturally occurring claims, we aim to train a generative model that encodes PIO elements that then decodes a medical claim.

Once a fine-tuned generative model is availabl a medical evidence dataset is required to generate synthetic claims. Trialstreamer \citep{marshall_trialstreamer_2020} offers a dataset of all randomised control trials (RCTs) along with the PIO elements, extracted using machine learning methodologies \citep{marshall_automating_2017}, related to those publications. Once a generative model is fine-tuned using RedHOT to generate medical claims for a given PIO set, this means for every available piece of medical evidence, we can use the PIO span from that evidence as input to generate a synthetic medical claim resulting in an aligned claim-abstract pair.

In this study, we drop the use of the $C$ element, a comparator is always an intervention, and it is unnecessary to create a distinction in our generators and classifiers in the context of this task \citep{brockmeier_improving_2019, chabou_combination_2018}.

\section{Expansive Medical Claim Corpus}

\begin{figure*}[!ht]
    \centering
    \includegraphics[scale=0.325]{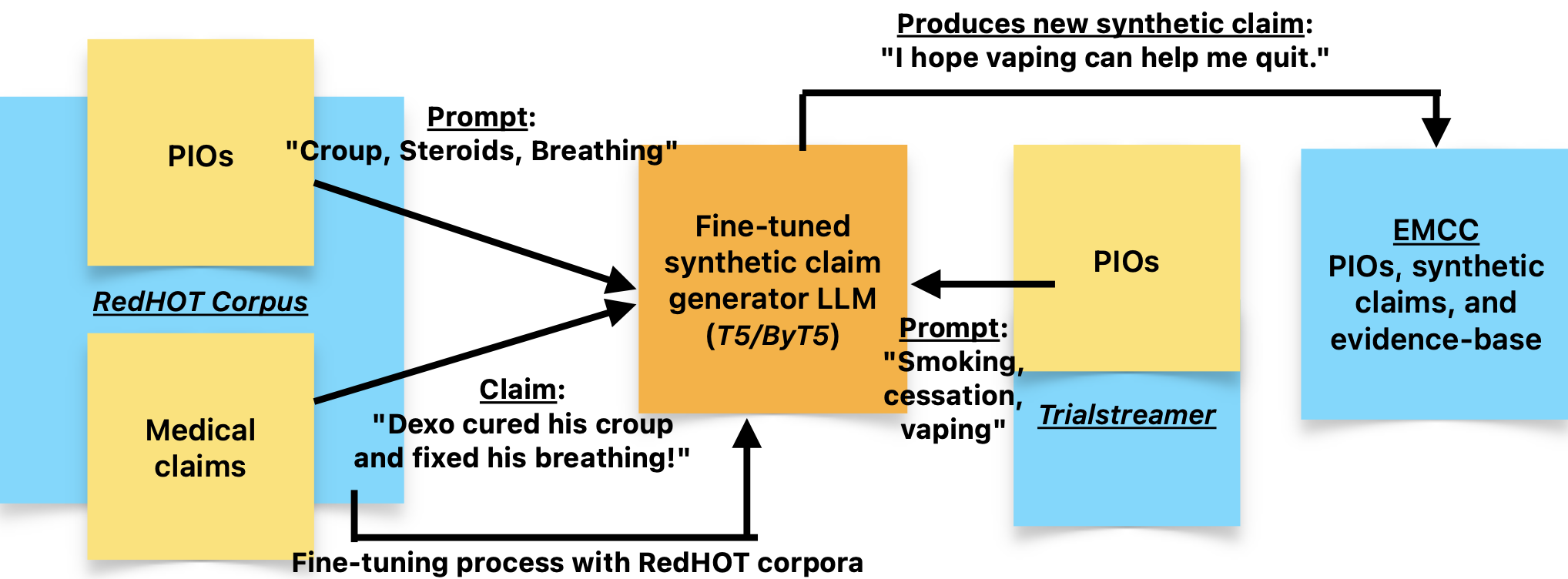}
    \caption{Figure displaying the flow of data for the curation of EMCC.}
    \label{fig:training-pipeline}
\end{figure*}

The methodology described in this section is for creating the Expansive Medical Claim Corpus (EMCC) where that methodology can be used to synthesise medical claims for any unseen PIO sequence. The corpus is generated with three steps: (1) curation of RedHot dataset, (2) using the selected dataset as the foundation for fine-tuning a pretrained generative model, and (3) observing the quality produced by the fine-tuned generative model through qualitative analysis. See \autoref{fig:training-pipeline} for an overview of the synthetic generator methodology.

The basis of the training data was curated by taking all the of the medical claims in the RedHOT corpora, and the PIO elements that they are annotated with. The PIO elements are fed to the encoder and the medical claim fed to the decoder of a given generative model. The aim with this methodology is that a model will learn to generate naturally sounding claims given an unseen set of PIO elements.

Three models were selected in this work for generating the synthetic medical claims: Falcon-7B \citep{penedo_refinedweb_2023}, T5 \citep{raffel_exploring_2019} and ByT5 \citep{xue_byt5_2022}. Due to computational power constraints, the majority of our experiments utilise \textit{t5-base} and \textit{byt5-base}.

Falcon is a seven billion parameter autoregressive decoder-only model trained on trillions of tokens, and has demonstrated high performance in a wide range of language tasks. We utilise this model because it will allow us to draw comparisons between a large scale model versus small models, such as the T5 family. T5 is capable of performing any text generation task, and this capability makes T5 flexible in a wide variety of tasks without task-specific architectures. For example, T5 has shown promise in the generation of synthetic corpora for QA tasks \citep{agarwal_knowledge_2021}. We conduct further experiments with ByT5, a byte-level encoder-decoder model, and variant of T5. By altering the tokenisation technique to operate on only the bytes within a string, as opposed to the words or subwords, the tokenisation is potentially more robustness against noise and more capable in multilingual settings. This token free approach has shown promising results in tasks with noisy data \citep{samuel_ufal_2021}, and further results shown in its ability to generate synthetic data \citep{stankevicius_correcting_2022}.

The selected models are then further fine-tuned with the curated dataset. The curation process takes all of the medical claims in the RedHOT corpora that are annotated with the PIO elements, as per the example shown in \autoref{tab:table_1}. The data is to be split into input (PIO elements) and target text (medical claim) for the encoder and decoder respectively. We split the curated dataset into training (90\%) and validation (10\%) for qualitative analysis. A specific test set was not isolated at this stage as generation quality was monitored qualitatively post training. The aim of this methodology is that the model will generate naturally sounding claims given an unseen set of PIO elements from the RedHOT corpus.


A qualitative analysis takes place at the end of each training cycle to observe the synthetic claims being generated by the model. This process requires randomly selecting PIO elements from abstracts within Trialstreamer, unseen from within the RedHOT corpus, and then using the generative model to generate medical claims. These tests are all PIO sets that were previously unseen to the model to understand model adaptability to unseen questions.

Finally, we focus on two decoding strategies: contrastive search \citep{su_contrastive_2022} and multinomial sampling \citep{keskar_ctrl_2019}. We utilise these strategies due to their results in handling repetition and tasks regarding fluency.

\section{Implementation Detail}

In this section, we describe the details of the EMCC's generation, including the qualitative-only analysis and hyperparameters.

\subsection{Hyperparameters}

The following hyperparameter values were manually selected: batch size $1$, maximum epochs $20$ with an early stopping mechanism, AdamW optimizer and learning rate $1e-4$ for both classification tasks - we use identical hyperparameters for our two selected models, however the sequence lengths differ. These are important to the selected models and the task of generating medical claims, because claims are generally limited to single sentences and short utterances. We, therefore, alter the sequence lengths to reflect this idiosyncrasy. The minimum and maximum sequence lengths differ substantially between the character-level model, ByT5, and the token-level model - T5. Given that ByT5 operates at the byte-level it means that a single byte is generated for each element of a sequence, where as token models generate a token which is made up of many bytes. ByT5 was given a minimum length of $56$ and a maximum length of $256$, in contrast to the the token-level model where we selected $28$ and $128$.

Once training had begun with the prepared data, the training loss was observed at each epoch until it had stabilised and then training was halted.



\subsection{Qualitative-Only Analysis}
\label{eval:decoding}

For this study, we performed a qualitative-only analysis of our synthetic medical claims. With this we took an iterative approach to analyse a test set of PIO claims unseen in the training data. 

Different prompts containing PIO elements, taken at random from the medical evidence-base, Trialstreamer, were used to observe the synthetic medical claims being generated. First observations show that models have potential for generating naturally sounding medical claims. We looked to identify the following criteria: claims appear to be coherent, claims appear natural, and if were being read by a patient could benefit from being aligned with medical evidence. See \autoref{tab:synthetic-qual-smoking} for an example.

\begin{table}[!ht]
\begin{center}
\begin{tabular}{ c|c }
 \hline
 \multicolumn{2}{c}{\textbf{Prompt}: \textit{smokers, nicotine, cessation}} \\
 \hline
 \textbf{\textit{Model}} & 
    \textbf{\textit{Output}} \\
 \hline 
  \textit{ByT5} 
    & \multicolumn{1}{p{5.5cm}}{I know that smokers are not an oral medication for over 14 days} \\
 \textit{T5} 
    & \multicolumn{1}{p{5.5cm}}{they often talk about how they feel about nicotine and smoking cessation...} \\
 \textit{Falcon} 
    & \multicolumn{1}{p{5.5cm}}{I was told the cessation may have long term effects on reduced mortality....} \\
 \hline
\end{tabular}
\caption{Table displaying an example prompt and the synthetic claims.}
\label{tab:synthetic-qual-smoking}
\end{center}
\end{table}

In \autoref{tab:synthetic-qual-kidney}, we observe a problem in the synthetic claims being generated by the fine-tuned models. Although these sequences appear natural, they are either vague or not a claim. The response by T5, in \autoref{tab:synthetic-qual-kidney}, identifies a personal experience rather than a specific claim. Using the same prompt, ByT5 generates a natural sounding claim, however due to its vagueness it becomes difficult to assess how relevant it is to the PIO used in the prompt.

\begin{table}[H]
\begin{center}
\begin{tabular}{ c|c }
 \hline
 \multicolumn{2}{c}{\textbf{Prompt}: \textit{Transplant, Immunosuppressant, }} \\
 \multicolumn{2}{c}{\textit{Diabetes Mellitus}} \\
 \hline
 \textbf{\textit{Model}} & 
    \textbf{\textit{Output}} \\
 \hline 
  \textit{ByT5} 
    & \multicolumn{1}{p{5.5cm}}{I've read online how the reactions are possibly caused by drug restrictive abnormalities.} \\
 \textit{T5} 
    & \multicolumn{1}{p{5.5cm}}{I got a free Kidnet transplant for Diabetes Mellitus} \\
 \textit{Falcon} 
    & \multicolumn{1}{p{5.5cm}}{I got
my first Immunosuppressant for my Diabetes Mellitus in May.} \\
 \hline
\end{tabular}
\caption{Table displaying an example prompt and the synthetic claim.}
\label{tab:synthetic-qual-kidney}
\end{center}
\end{table}

In the example \autoref{tab:synthetic-strategy-smoking}, we see naturally sounding synthetic claims, however the constrastive search is producing more claims that adhere to the quality criteria we specified.

\begin{table}[H]
\begin{center}
\begin{tabular}{ c|c }
 \hline
 \multicolumn{2}{c}{\textbf{Prompt}: \textit{smokers, nicotine}} \\
 \multicolumn{2}{c}{\textit{smoking cessation}} \\
 \hline
 \textbf{\textit{Strategy}} & 
    \textbf{\textit{Output}} \\
 \hline 
  \textit{T5} \& \textit{Contrastive} 
    & \multicolumn{1}{p{3.5cm}}{Some people say it's better to quit smoking when their symptoms change} \\
 \textit{ByT5} \& \textit{Contrastive} 
    & \multicolumn{1}{p{3.5cm}}{I know there are some good stories of smokers having successful cessation of smoking.} \\
 \hline
  \textit{T5} \& \textit{Multinomial} 
    & \multicolumn{1}{p{3.5cm}}{I know smokers are also strong enough to cause smoking cessation in the marginal smokers} \\
 \textit{ByT5} \& \textit{Multinomial} 
    & \multicolumn{1}{p{3.5cm}}{I have heard that smokers flare ups} \\
 \hline
\end{tabular}
\caption{Table displaying a contrast between the two decoding strategies. The prompt is a PIO span, and the output is the synthetic claim.}
\label{tab:synthetic-strategy-smoking}
\end{center}
\end{table}

Experiments using Falcon were dropped at this stage, in order to have high speed inference it requires significant compute resources, and it’s qualitative analysis did not suggest that it was generating claims that were any more naturally
sounding than that of the smaller models.

After repeating this process with other sets of PIO elements, we identified at this point that the constrastive model was generating more natural sounding claims, particularly when used with T5. We next went on to experiment with other hyperparameters, such as temperature, to understand the impacts on the sequence being generated by both models. ByT5 was often repeating itself, therefore we iteratively altered our decoding parameters. In the appendixes we attach additional information; see \autoref{tab:decoding-params} for the final selected decoding parameters , and \autoref{tab:temperature-strategy} for
more details on decoding experiments in the appendix.

Additionally, we also conducted small scale one- and few-shot experiments using the GPT-3 model series, we have attached these results in the appendices. See \autoref{appendix:gpt35-experiments} for additional information.

In the appendix, \autoref{appendix:stats} and \autoref{tab:corpus-stats}, we share an overview of the final available corpora generated using T5 and the discussed decoding parameters.


\section{Downstream Task Evaluation}

We evaluate the quality of our generated corpus (EMCC) using two downstream tasks Evidence retrieval (Section~\ref{sec:eviretri}) and Claim and PIO Identification (Section~\ref{sec:bioidentification})

\begin{table*}[t]
\begin{center}
\begin{tabular}{ c|c|c|c|c|c|c|c|c|c|c|c|c|c }
 \hline
  \textit{.} & \textit{I} & \textit{had} &  \textit{lupus} & \textit{erythematosus} & 
  \textit{for} & \textit{12} & 
  \textit{years} & \textit{the} & \textit{pain} & \textit{was} & \textit{bad} & \textit{.} & \textit{We} \\
  \hline
  O & B & I & I & I & I & I & I & I & I & I & I & I & O \\
\hline
\end{tabular}
\caption{Table showing an excerpt of a social media post annotated with the BIO classes to identify the medical claim.}
\label{tab:span-example}
\end{center}
\end{table*}

\subsection{Claim and PICO Identification}
\label{sec:bioidentification}
Given a social media post as a sequence of text, the task is to annotate that post with the necessary labels in order to identify the medical claims within it and further identify any PIO elements. As proposed by \cite{brockmeier_improving_2019}, we use a Beginning-Inside-Out (BIO) style format to label \textbf{B}eginning of the PICO mention, \textbf{I}nside of the PICO mention, and \textbf{O}ut side of the PICO mention. See \autoref{tab:span-example} for a full example. We perform a similar labeling approach to that of PICO identification. In this task, we utilise each of the the BIO classes in conjunction with each of the PIO categories: $Pop$ represents Population, $Int$ represents Intervention and $Out$ represents Outcome. This can be seen in the example in \autoref{tab:pico-span-example}.

\begin{table*}[h]
\begin{center}
\begin{tabular}{ c|c|c|c|c|c|c|c|c|c|c|c|c|c }
  \hline
  \textit{.} & \textit{I} & \textit{had} &  \textit{lupus} & \textit{erythematos} & 
  \textit{,} & \textit{1} & 
  \textit{year} & \textit{the} & \textit{pain} & \textit{was} & \textit{bad} & \textit{.} & \textit{We} \\
  \hline
  O & O & O & B-Pop & I-Pop & O & O & O & O & B-Out & O & O & O & O \\
\hline
\end{tabular}
\caption{Table showing an example of a medical claim annotated with the population and outcome classes to identify the PIO spans.}
\label{tab:pico-span-example}
\end{center}
\end{table*}


Two models were selected for this task; a statistical machine learning model and a deep learning model. Conditional Random Fields (CRF) was chosen for building a baseline, as is often reflected in the literature \citep{chabou_combination_2018, kim_automatic_2011}, and an encoder-only transformer-based model was used for comparison with current state-of-the-art models \citep{wadhwa_redhot_2023}.

\textbf{Statistical Model} We extract and use similar features to that of \citet{wadhwa_redhot_2023} to train our model, each token is labelled with the following feature: part-of-speech (POS) tag, it is alphanumeric, it is an uppercase word, it is numeric, it is title-case, the previous three words and their POS tags, the following three words and their POS tags.

\textbf{Transformer Model} We look to use BERT-based uncased, and also adapt this model on the new synthetic claims. We hypothesise that by adapting our encoder-only model on our new in-domain synthetic corpora and further fine-tuning on the task specific dataset, when testing the model on that existing RedHOT corpora, we can address our original research question.

Given that our synthetic dataset is not an annotated corpus of social media like that of RedHOT, we require a different approach to training as we can't fine-tune specifically for span classification within posts. We look to reuse the BERT masked language modelling strategy for in-domain adaptation, further fine-tune and then evaluate using the RedHOT corpora.


The following hyperparameter values were manually selected: batch size $8$, maximum epochs $20$ with an early stopping mechanism, AdamW optimizer and learning rate $1e-5$ for both classification tasks.

For both experiments we used an uncased base BERT model and also the domain adapted BERT trained using masked language modelling on the synthetically generated medical claims. We selected $15\%$ whole word token masking approach to our corpus. Once the corpus was sufficiently masked, the following hyperparameter values were manually selected: batch size $32$, an optimizer with a learning rate $2e-5$ and 1000 warm-up steps.

\subsubsection{Results}

\begin{table}[h]
\begin{center}
\begin{tabular}{ c|c|c|c } 
\hline
  \textbf{\textit{Model}} & \textbf{\textit{Precision}} & \textbf{\textit{Recall}} & \textbf{\textit{F1}} \\
 \hline 
 \textbf{CRF} & 70.24 & 58.37 & 63.76 \\
 \textbf{BERT} & 52.63 & 58.82 & 47.61 \\
 \textbf{BERT-Synth} & \textbf{72.24} & \textbf{68.30} & \textbf{71.15} \\
 \hline
\end{tabular}
\caption{Table displaying the results of our claim classification models on the RedHOT \citep{wadhwa_redhot_2023} test set. BERT result obtained from \citet{wadhwa_redhot_2023}.}
\label{results:pico-claim-redhot}
\end{center}
\end{table}

In \autoref{results:pico-claim-redhot}, we display the results of our claim classification performance on a test set derived from the original RedHOT corpus \citep{wadhwa_redhot_2023}.


In \autoref{results:pico-class-redhot}, we display the results of our claim classification performance on a test set derived from the original RedHOT corpus \citep{wadhwa_redhot_2023}.

We find in our results that BERT adapted to our synthetic dataset, then tested on the RedHOT corpus is the highest performing system by 7\%, where CRF is the second most performant. This beats all current state-of-the-art methods \cite{wadhwa_redhot_2023, ramachandran_masonnlp_2023}.

\begin{table}[]
\begin{center}
\begin{tabular}{ c|c|c|c } 
 \hline
  \textbf{\textit{Model}} & \textbf{\textit{Precision}} & \textbf{\textit{Recall}} & \textbf{\textit{F1}} \\
 \hline
 \textbf{CRF} & 32.42 & 32.42 & 32.42 \\
 \textbf{BERT} & 43.88 & 36.13 & 39.62 \\
 \textbf{BERT-Synth} & \textbf{49.45} & \textbf{41.27} & \textbf{44.97} \\
 \hline
\end{tabular}
\caption{Table displaying the results of our PIO classification models on the RedHOT \citep{wadhwa_redhot_2023} test set. BERT result obtained from \citet{wadhwa_redhot_2023}}
\label{results:pico-class-redhot}
\end{center}
\end{table}

We find in our results that BERT adapted to our synthetic dataset, fine-tuned and tested on the RedHOT corpus is the highest performing system by 3\%, where BERT without any further training on the synthetic data is the second most performant. This beats all current state-of-the-art methods \cite{wadhwa_redhot_2023, ramachandran_masonnlp_2023}.

\subsubsection{Discussion}

Our results for claim classification suggest that a statistical model is far more performant than a BERT model that is not domain adapted. CRF has a 16 point improvement over BERT, however when BERT is adapted to the domain there was an increase of 7 points over the statistical model. We note that our CRF system has a 25 point improvement over the original works \citep{wadhwa_redhot_2023}, this is likely due to our methodology containing more features. This heavy feature engineering process, however, is not required by deep learning models making them more adaptable to change.

We find for PIO classification that utilising BERT is more performant than the statistical model. We also find that model \textbf{BERT-Synth} has a 2\% increase in F1 score than the existing works \citep{wadhwa_redhot_2023}. This is suggestive that domain-specific synthetic data can have a positive impact on PIO classification, furthermore this was performed on a smaller model than that of the state-of-the-art.

We have been able to demonstrate that deep learning mechanisms vastly out perform statistical models for the tasks within this study. Given the expansive body of literature in deep learning models we expected an improvement using an encoder-only architecture, however a unique improvement we also see is the introduction of PIO elements into the query. The addition of this into the query elements offers a significant performance improvement across all model architectures.

Given that our results are centered around synthetic medical claims and therefore synthetically aligned documents, we felt that this bias could be alleviated using a medical expert to perform manual evaluation against real medical claims retrieved from social media. In the results, we found it was clear that the system is capable of returning medical abstracts that are relevant to the claim. Moreover, in some instances, the system is capable of returning highly relevant documents. These abstracts offer tangible advice that could utilised by a patient of professional in order to resolve or investigate the claim further.

\subsection{Evidence retrieval}
\label{sec:eviretri}

Evidence retrieval can be framed as a question-answering task, whereby we use the PIO elements and the medical claim as the question and the answer is a set of appropriate medical abstracts.

The aim of the evidence retrieval component is to retrieve the most pertinent evidence in relation to a given medical claim and it's associated PIO data. This requires being able to evaluate the accuracy of the system in regards to its pertinence, therefore we look to use the metric precision@$k$ \citep{jirvelin_ir_2017} in order to do this. $k$ represents the quantity of documents being returned by the system, and we look to count how many relevant documents are found within that specified quantity of results.

We utilise two machine learning methodologies for this task: a statistical model and a deep learning model. BM25 \citep{robertson_probabilistic_2009} is utilised for deriving a baseline as seen in the literature \citep{wadhwa_redhot_2023} and dense passage retrieval (DPR), a deep learning technique is used to observe performance against the existing state of the art \citep{wadhwa_redhot_2023}.

BM25 and DPR requires separating queries from answers, therefore we prepare our dataset, as such that our synthetic claims and PIO elements are separate from the medical abstracts. We propose the use of two experiments: (1) is where only the synthetic claim is used as a query to fetch medical abstracts, and (2) is where the concatenation of claim, and PIO elements are used as a query to fetch medical abstracts.

The literature \citep{karpukhin_dense_2020} notes that DPR is highly resource intensive, therefore we will only experiment with the query construction that is the highest performing from the BM25 experiments. Due to the time and resources required to train a resource intensive model, we leave more intensive experiments for future work.

Performance in DPR is greater when negative samples \citep{gillick_learning_2019} are provided along with the positive samples. We construct a negative sample set of abstracts for every claim-evidence pair where those PIO elements are not used. This would mean for the PIO elements "\textit{lupus, methotrexate, plaquenil nauseous}", we would take all other aligned PIO claim abstracts for our negative set.

\subsubsection{Results}

We believe it to be necessary to conduct evidence retrieval experiments where the query is only the synthetic medical claim, and a second set of experiments where the PIO elements and the synthetic claim are concatenated.

We report the results of the evidence retrieval task in \autoref{results:ev-ret}. We conducted three experiments over two models, where both models were tested with PIO elements concatenated with claims as queries, also a claim only query experiment was conducted.

\begin{table}[H]
\begin{center}
\begin{tabular}{ cc|c|c|c }  
 \hline
 \multicolumn{5}{c}{\textbf{Precision @ \textit{k}}} \\
 \hline
 \textbf{\textit{k}} & 1 & 5 & 10 & 100 \\
 \hline
 BM25-Claim & 1.5 & 5.1 & 7.8 & 17 \\
 \hline
 BM25-PIO & 4.1 & 12.8 & 17.3 & 35.5 \\
 \textbf{DPR-PIO} & \textbf{8.17} & \textbf{15.49} & \textbf{24.72} & \textbf{36.80} \\
 \hline
\end{tabular}
\caption{Table displaying the results of all automated evidence retrieval experiments.}
\label{results:ev-ret}
\end{center}
\end{table}

As seen in \autoref{results:ev-ret}, we first find that where the evidence query contains PIO and claim elements, \textit{\textbf{BM25-PIO}} and \textit{\textbf{DPR-PIO}}, there is a significant improvement in precision. \textit{\textbf{DPR-PIO}} gives an 8 times improvement in results when $k$ is set to 1, and at least double in all other $k$ scoring experiments.

\subsubsection{Expert Evaluation}

To further validate the evidence retrieval system, we look to utilise the naturally occurring claims from the RedHOT corpora and query our trained evidence retrieval system with those claims. An expert, in the form of a qualified doctor currently practising as a first year general practitioner (GP), is utilised to perform the analysis.

We first randomly select 15 medical claims from the RedHOT corpus. These are then presented to the Doctor with the first 5 medical abstracts returned from our highest performing system (DPR-PICO) in ranked order. We then asked the expert to allocate a score to each of the 5 abstracts. The score was based on the works of \cite{wadhwa_redhot_2023}, where the expert was asked to allocate a relevant, somewhat relevant or irrelevant label to each abstract. We expand on this evaluation, on advice from the medical expert, with an additional scoring category - highly relevant. We introduce a highly relevant category to identify abstracts that provide direct and tangible advice that an expert or non-expert could utilise in a real-world setting. These scores can be used within a precision@$k$ style scoring to understand the relevancy of abstracts at different precision counts
.  

\begin{table}[H]
\begin{center}
\begin{tabular}{ cc|c }  
 \hline
 \multicolumn{3}{c}{\textbf{Precision@\textit{k}}} \\
 \hline
 \textbf{\textit{k}} & 1 & 5 \\
 \hline
 Highly Relevant & 0 & 3 \\
 Relevant & 3 & 13 \\
 Somewhat Relevant & 5 & 23 \\
 Irrelevant & 7 & 36 \\ 
 \hline
\end{tabular}
\caption{Table displaying the results of the manual expert analysis of the DPR-PICO evidence retrieval system.}
\label{results:expert}
\end{center}
\end{table}

In \autoref{results:expert}, we display our results in regards to the expert evaluation. We find that when the precision@$k$ is higher more irrelevant results are introduced, however the total sum of each relevant category always outweighs that of the irrelevant category.

\section{Limitations}

A limitation of this work is that we were vastly limited on how many data points we could review from our synthetic corpus. It is, therefore, possible that noise or poorly generated claims are stored within the dataset. An alternative to the qualitative approach approach may be to build a classifier that dictates whether or not a generated claim is acceptable or not. This acceptability is decided by a classifier trained upon a dataset of naturally occurring claims, i.e the data from \cite{wadhwa_redhot_2023}. A further improvement for this would be to domain-adapt the encoders in the evidence retrieval architecture as we did for the classification results.

We felt we were limited in our ability to perform a like-for-like comparison as previous works did not share the claims and evidence they used to perform their expert evaluation. Due to limited resources too, we were unable to a test a similar quantity of claims and abstracts. We leave these issues for future works.

\section{Conclusions}

\textbf{RQ1} addressed whether a methodology could be constructed to produce a synthetic corpus of medical claims, and whether that corpus could improve downstream tasks. A requirement of this corpus is that it be aligned with PIO elements and medical abstracts. We believe we constructed a working methodology that is capable of generating a synthetic claim for any given PIO structure from the medical literature. This corpora was qualitatively-only evaluated, and in future we look for automated quantitative measures of quality.

\textbf{RQ2} addressed to what extent a classification system could identify medical claims in social media posts. We conducted experiments to look at how well system perform on existing corpora on both statistical and deep learning architectures. We found that deep learning architectures gave us a $7\%$ increase in performance, when utilising EMCC in an in-domain adaptation technique using the new synthetic corpora on a base large language model.

\textbf{RQ3} addressed to what extent a classification system could identify medical claims in social media posts. We conducted experiments to observe the performance on existing corpora and synthetic corpora. We found that deep learning architectures adapted on the EMCC gave us a $12\%$ increase over statistical methodologies.

\textbf{RQ4} addressed how precisely we can retrieve medical evidence that relates to medical claims made in social media posts. We conducted our experiments using the synthetic dataset, where each claim in that set is paired with a medical abstract.

In our results, we saw that a deep learning architecture vastly outperforms that of the statistical models. Furthermore, the introduction of additional context in the query, in this task it was the PIO elements, improves the precision by 3 times when precision@$k$ is set to one. In our expert analysis, we find that for almost every query we get a least $1$ relevant document when precision@$k$ is set to five.



\section{Acknowledgements}
This work was supported by the Centre for Doctoral Training in Speech and Language Technologies (SLT) and their Applications funded by UK Research and Innovation (Grant number EP/S023062/1).






\nocite{*}
\section{Bibliographical References}\label{sec:reference}

\bibliographystyle{lrec-coling2024-natbib}
\bibliography{references}

\section{Language Resource References}
\label{lr:ref}
\bibliographystylelanguageresource{lrec-coling2024-natbib.bst}
\bibliographylanguageresource{languageresource}
Anthony Hughes and Xingyi Song. 2023. Expansive Medical Claim 
Corpus. EMCC is distributed via Zenodo.
\url{https://zenodo.org/records/8321460}.
DOI 0.5281/zenodo.8321460

\section{Appendices}

\subsection{Dataset Statistics}
\label{appendix:stats}

In \autoref{tab:corpus-stats}, we share an overview of our available corpora.

\begin{table}[h]
    \centering
    \begin{tabular}{l|c}
     \hline
         \textbf{\textit{Property}} & \textbf{\textit{Count}} \\
         \hline
         Claims & 129363 \\
         Sentences & 129363 \\
         All Tokens & 2769794 \\
         Unique Tokens & 56824 \\
         Unique PIOs & 84702  \\
         Count of all populations & 8948 \\
         Count of all interventions & 10855 \\
         Count of all outcomes & 8445 \\
         \hline
    \end{tabular}
    \caption{Table displaying properties and counts of the new synthetic claims corpus.}
    \label{tab:corpus-stats}
\end{table}

\subsection{GPT-3.5 Experiments}
\label{appendix:gpt35-experiments}

The following section looks at the small scale experiments we conducted to investigate the plausibility of using an LLM as a means of generating synthetic claims. We used document examples manually selected from the medical literature and augmented these into prompts.

\subsubsection{GPT-3.5 In-Context Zero-Shot Experiments}
\label{appendix:gpt3-oneshot-experiments}

In our in-context zero-shot experiments, we utilised ChatGPT without fine-tuning, however providing the LLM with an example within the context window of the prompt. \autoref{tab:gpt3-oneshot-outputs} displays an example prompt and response we provided from the GPT-3.5-turbo model.

\begin{table*}[]
\begin{center}
\begin{tabular}{l}
 \hline
 \textbf{PICO}\\
 \hline
    Populations: Coronary Artery Disease \\
    Interventions: menaquinone 7,Vitamin K antagonists \\
    Outcomes: Scores (qualifier value),Progression (attribute), Vascular Calcinosis \\
    Using this information write a short personal medical claim written by a reddit or twitter user. \\
    Include alternative names for the medical concepts. \\
    Include spelling and grammatical mistakes, and use colloquial language. \\
 \hline
 \textbf{Response}\\
 \hline
I'm really curious about this VitaK-CAC trial!\\
They're testing out if Vitamin K can help slow down arterial calcification for patients with \\ 
Coronary Arteries Disease.\\
I heard it can even help with Vascular Necrosis.\\
\hline
\end{tabular}
\caption{Table displaying a prompt provided to GPT-3.5 and the response returned from the model.}
\label{tab:gpt3-oneshot-outputs}
\end{center}
\end{table*}

\subsubsection{GPT-3.5 Few-Shot Experiments}
\label{appendix:gpt3-fewshot-experiments}

In our in-context example, we felt the responses were of sufficient enough quality that using an LLM for generating synthetic claims warranted further investigation. In \autoref{tab:gpt3-fewshot-training} we share a set of example documents that we provided to gpt-3.5-turbo during fine-tuning. The set of example documents were taken from \cite{wadhwa_redhot_2023}; a set of documents were manually selected, and then aligned PIO elements and medical claims isolated. Additionally, we provide document examples to demonstrate the outputs of this fine-tuning in \autoref{tab:gpt3-fewshot-outputs}.

\begin{table*}[]
\begin{center}
\begin{tabular}{l|c}
 \hline
 \textbf{Prompt} & \textbf{Example Response}\\
 \hline
 allo, uric acid & 
 I've been on allo for the past 18 months and have had no flares \\
 \hline
 gaps diet,gerd & I've read that the gaps diet can heal things like gerd \\
 \hline
 inflammation, gout, flare ups & I got my first attack in February \\ & and was following the advice of avoiding high purine foods. \\
\hline
\end{tabular}
    \caption{Table displaying a set of examples provided to GPT-3.5 and the ground-truth response the model is required to reproduce.}
\label{tab:gpt3-fewshot-training}
\end{center}
\end{table*}

\begin{table*}[]
\begin{center}
\begin{tabular}{l|c}
 \hline
 \textbf{Prompt} & \textbf{Fine-tuned Model Response}\\
 \hline
 marijuana,IBS,cancer & 
 Because I heard it helps with certain medical conditions. \\
 & I know it can have psychoactive effects too \\
 \hline
 asthma, albuterol, hospitalization &  I try to minimize albuterol use to days I feel particularly bad \\ & because I know albuterol can only help so much \\
 \hline
  hyperglycemia,nutrition & Ive seen a few studies on the effects of sub-Q injections in \\ & people with Type 1 Diabetes experiencing hyperglycemia \\
\hline
\end{tabular}
    \caption{Table displaying a set of prompts provided to a fine-tuned GPT-3.5 and the response of the model.}
\label{tab:gpt3-fewshot-outputs}
\end{center}
\end{table*}

\begin{table*}[]
\begin{center}
\begin{tabular}{ c|c|c|c|c|c|c }
 \hline
 \textbf{\textit{Model}} &
    \multicolumn{1}{p{2cm}|}{\textbf{\textit{No. of Beams}}} &
    \multicolumn{1}{p{2cm}|}{\textbf{\textit{Min. Seq. Length}}} &
    \multicolumn{1}{p{2cm}|}{\textbf{\textit{Max Seq. Length}}} & 
    \textbf{\textit{Temp}}. & 
    \multicolumn{1}{p{2cm}|}{\textbf{\textit{Penalties}}} & 
    \multicolumn{1}{p{2cm}}{\textbf{\textit{No Repeat nGram Size}}} \\
 \hline
 ByT5 & 1 & 56 & 256 & 0.5 & 0.5 & 15 \\
 T5 & 1 & 28 & 84 & 0.8 & 0.5 & 3 \\
 Falcon & 1 & 28 & 128 & 0.5 & 0.5 & 3 \\
 \hline
\end{tabular}
\caption{Table displaying a summary of the final parameters utilised to find an optimal generation strategy. The number of beams applied to the output of the decoder as well as the sequence length, and controlling any issues with repetition all formed part of the decoding strategy. We selected T5 for the creation of the EMCC.}
\label{tab:decoding-params}
\end{center}
\end{table*}

\begin{table*}[]
\begin{center}
\begin{tabular}{ c|c|c }
 \hline
 \multicolumn{3}{c}{\textbf{Prompt}: \textit{smokers and nicotine and smoking cessation}} \\
 \hline
 \textbf{\textit{Strategy}} & 
    \textbf{\textit{Temperature}} & 
    \textbf{\textit{Output}} \\
 \hline 
  \textit{T5} \& \textit{Contrastive} 
    & 0.2
    & \multicolumn{1}{p{8cm}}{they often talk about how they feel about nicotine and smoking cessation.} \\
 \textit{ByT5} \& \textit{Contrastive} 
    & 0.2
    & \multicolumn{1}{p{8cm}}{they often talk about how they feel about nicotine and smoking cessation} \\
 \hline
   \textit{T5} \& \textit{Contrastive} 
    & 0.5
    & \multicolumn{1}{p{8cm}}{Some even say it stops them from smoking.} \\
 \textit{ByT5} \& \textit{Contrastive} 
    & 0.5
    & \multicolumn{1}{p{8cm}}{I know that it can cause smoking cessation} \\
 \hline
   \textit{T5} \& \textit{Contrastive} 
    & 0.8
    & \multicolumn{1}{p{8cm}}{I know there are some good stories of smokers having successful cessation of smoking. Some even say the smoking cessation is a good idea.} \\
 \textit{ByT5} \& \textit{Contrastive} 
    & 0.8
    & \multicolumn{1}{p{8cm}}{I know there is no guarantee going off smoking cessation.} \\
 \hline
\end{tabular}
\caption{Table displaying a contrast between the two decoding strategies. The example prompt is a PIO span, and the output is the synthetic claims being generated by the model.}
\label{tab:temperature-strategy}
\end{center}
\end{table*}

\end{document}